\documentclass[journal]{IEEEtran}

\usepackage[T1]{fontenc}
\usepackage{newtxtext,newtxmath}
\usepackage{amsmath}
\usepackage{graphicx}
\usepackage{booktabs}
\usepackage{tabularx}
\usepackage{array}
\usepackage{xcolor}
\usepackage{enumitem}
\usepackage{tikz}
\usetikzlibrary{arrows.meta,positioning,fit,calc}
\usepackage[hidelinks]{hyperref}

\graphicspath{{figures/}}
\newcolumntype{Y}{>{\raggedright\arraybackslash}X}
\newcommand{\Eclass}{\mathcal{E}}
\newcommand{\Hpool}{\mathcal{H}}
\newcommand{\Ouser}{\mathcal{O}_u}

\title{From Generation to Matching: A Development Report on Personalized Chinese Handwriting}

\author{Yiwei Liu\\
The Chinese University of Hong Kong, Shenzhen\\
\href{mailto:226015067@link.cuhk.edu.cn}{226015067@link.cuhk.edu.cn}}

\begin{document}

\maketitle

\begin{abstract}
This paper documents a frozen engineering project on personalized Chinese handwriting. The project started from approximately 200 real handwriting images from one user, covering 197 unique Chinese characters, and was initially formulated as few-shot generation of unseen characters. A sequence of canonical-centered personalization routes repeatedly exposed the same conflict: increasing structural pressure made outputs more canonical, while increasing personalization could damage identity-defining strokes. The project was therefore reset around real-human character equivalence classes. A multi-writer CASIA candidate pool showed that a USER-compatible realization often already existed among valid human samples. The task consequently changed from synthesis to character-wise matching, followed by cross-writer composition into a virtual writer. The frozen system uses real-ink features, character-specific human population percentiles, top-20 candidate pruning, and greedy hardest-first whole-row selection. On the covered target set, all 197 USER characters had real-human candidates, and the 100-character evaluation subset was covered 100/100. Known-character held-out comparisons included a row judged visually almost indistinguishable from genuine USER handwriting. A 60-episode stability audit placed every episode in a predefined A-like machine-proxy region, but these were not independent human A-level judgments. The final evidence supports stable practical B-level quality, with many outputs approaching A-level under the USER-defined criterion. The report records why generation became unnecessary for this case without claiming unrestricted or universal handwriting synthesis.
\end{abstract}


\section{Problem}

The concrete input to this project was small and personal. Let
\begin{equation}
  \Ouser = \{(c_i,x_{u,c_i})\}_{i=1}^{m},
  \label{eq:observations}
\end{equation}
where $c_i$ is a Chinese character and $x_{u,c_i}$ is a real image written by the target user. The available USER set contained 200 verified handwriting images and 197 unique characters. The target was to obtain a glyph for a character not used during construction that still looked as if it could naturally belong to the same user.

Two requirements were present from the beginning. First, the output had to be the correct character. We call this character correctness. Second, the output had to occupy a plausible part of the target user's handwriting system. We call this writer likeness. They are related but not interchangeable:
\begin{equation}
  \text{character correctness} \neq \text{writer likeness}.
  \label{eq:two_axes}
\end{equation}

The natural first formulation was a few-shot generation problem,
\begin{equation}
  \Ouser \longrightarrow P_u(x\mid c),
  \label{eq:generation}
\end{equation}
where the observed USER glyphs should condition a distribution capable of producing a new realization for an unseen $c$. This formulation is intuitive when the desired output is described as ``the user's version of a character.'' It also makes the project look like a standard font- or style-generation problem. The development record shows why that intuition was incomplete.

The remainder of this report is organized by changes in the modeling assumption rather than by internal release chronology. It records failed routes as evidence, the final reformulation, and the limits of what the frozen system establishes.

\section{The Wrong Starting Point: Canonical-Font Personalization}

The early work did not fail because no method could produce a legible Chinese glyph. It failed because the object being personalized was given the wrong role. Several different families of methods converged on the same implicit geometry, and the failure modes became clearer when grouped by assumption.

\subsection{Canonical-centered assumption}

Diffusion personalization, style adaptation, writer embeddings, residual deformation, geometry transfer, and topology-preserving execution differed substantially in implementation. Their shared structure was closer to
\begin{equation}
  x_{\mathrm{standard}}(c) \longrightarrow x_{\mathrm{user}}(c)
  \label{eq:canonical_map}
\end{equation}
than to direct selection from a human handwriting population. A standard or printed glyph supplied content, topology, and often the visual coordinate system; personalization was asked to modify it without losing identity.

This is a reasonable engineering prior for some font-imitation settings. It is not a fact about how a person writes. The target user's sample for a character is one realization among many, and the canonical image is not known to be the center of that distribution. Giving it privileged status can make the output preserve the wrong invariants.

\subsection{Style--structure conflict}

A recurring hard case during development was the character recorded as U+653F. Some outputs acquired a convincing amount of USER-like slant, stroke weight, or local motion while losing a stroke or junction needed for the character to remain valid. Other outputs preserved the identity-defining structure but were visually close to a standard font. Stronger structural pressure moved results toward the canonical basin; stronger personalization increased the risk of missing or distorting character-defining strokes.

The important observation is not that one particular character is difficult. It is that a single scalar notion of ``similarity'' concealed two axes. An output can be style-positive but structure-invalid, or structure-correct but writer-negative. This became the main reason that later evaluations separated correctness from likeness rather than treating one as a proxy for the other.

\subsection{Global writer representations}

The next routes used writer tokens, latent embeddings, user adapters, source-invariant encoders, and preference or reranking models. These representations were useful for ranking some existing candidates. Some machine writer metrics also showed strong uplift. That uplift did not reliably survive direct USER inspection: a machine score could increase while the user still described the output as a variation of a standard font or as generic handwriting.

The limitation was coverage. A global writer code can select or emphasize a mode that already exists in a candidate set, but it cannot reliably create a USER-like realization for every character if the generator never reaches that region. The failure was therefore not solved by increasing the capacity of the writer representation.

\subsection{Local writing rules}

The project then measured reusable local signals across the observed characters. These included endpoint behavior, junction displacement, curvature, tangent direction, local extension, and local displacement. This was a real positive result: reusable cross-character local behavior existed. The same user did not write every character independently of all others.

However, estimating a local rule is not the same as safely executing a complete character. A rule that is helpful around an endpoint may be unsafe when applied to a different topology, stroke order, or junction context. Local predictability did not by itself provide a complete and valid unseen glyph.

\subsection{Topology-preserving execution}

Safety-oriented deformation and geometry methods made the constraint explicit. They protected connected components, stroke survival, or a topology proxy while transferring local variation. In the strongest cases, structural survival became very high. The same constraints also suppressed the variation that made the result feel personal. The outputs became canonical, over-regular, or clean in a way that was unlike the user's natural writing.

The resulting lesson was concise:
\begin{equation}
  \text{safe} \neq \text{user-like}.
  \label{eq:safe_not_like}
\end{equation}
Preserving an abstract structure is necessary for correctness, but it is not sufficient for writer likeness.

\subsection{Representation comparison and root-cause realization}

Global descriptors, local descriptors, morphology, topology, writer features, and hybrids were compared across the development record. The comparisons were useful because they removed several attractive explanations. Better rasterization did not remove the canonical appearance. Better writer scores did not guarantee human acceptance. More restrictive execution made the output safer but often less personal.

The most durable human feedback was that outputs remained variants of standard fonts. That feedback shifted the question from ``which representation should be added?'' to ``what should be represented?'' The problem was not simply a missing executor, embedding, or scorer. The canonical glyph itself might be the wrong modeling center.

\section{Character Identity as an Equivalence Class}

For a character $c$, define its legal realization equivalence class as
\begin{equation}
  \Eclass_c = \{x: x \text{ is a valid human realization of } c\}.
  \label{eq:equivalence}
\end{equation}
The class includes differences caused by font or writer, pen, pressure, slant, speed, and natural writing state. It is not a single shape with a small deformation neighborhood.

Both a standard glyph and a USER glyph can be members of this class,
\begin{equation}
  x_{\mathrm{standard}}(c)\in\Eclass_c,
  \qquad
  x_{\mathrm{user}}(c)\in\Eclass_c.
  \label{eq:members}
\end{equation}
The standard glyph is therefore an ordinary member, not a generation center. Human writing is not naturally performed by first imagining a canonical printed glyph and then deforming it. The mapping in~\eqref{eq:canonical_map} is an engineering hypothesis, not a consequence of character identity.

\subsection{Symbol and character equivalence as a diagnostic}

The symbol-equivalence diagnostic was valuable during this reset because it separated character identity from canonical similarity. It caught style-positive but structure-invalid cases, and the hard negatives remained useful regression cases. It helped make the U+653F failure concrete: a visually convincing output could still be the wrong character.

The later cross-domain real-human audit also exposed a strict limit. On 985 real-human pairs, the diagnostic produced only 2 PASS, 2 AMBIGUOUS, and 981 FAIL. This is severe false rejection when real CASIA handwriting is treated as input. The final system therefore uses trusted dataset provenance to establish character validity for real-human candidates. The diagnostic remains useful for analysis and regression, but it is diagnostic only and is not a hard veto. It should not be described as a robust cross-domain validator.

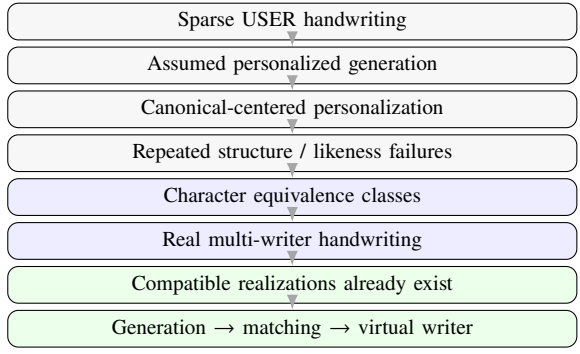
\begin{figure}[t]
\centering
\begin{tikzpicture}[x=1cm,y=1cm,>=Latex,
  stage/.style={draw,rounded corners,align=center,minimum width=0.85\columnwidth,minimum height=0.34cm,font=\footnotesize,fill=black!3},
  arrow/.style={->,semithick,gray!70}]
  \node[stage] (a) at (0,0) {Sparse USER handwriting};
  \node[stage] (b) at (0,-0.58) {Assumed personalized generation};
  \node[stage] (c) at (0,-1.16) {Canonical-centered personalization};
  \node[stage] (d) at (0,-1.74) {Repeated structure / likeness failures};
  \node[stage,fill=blue!7] (e) at (0,-2.32) {Character equivalence classes};
  \node[stage,fill=blue!7] (f) at (0,-2.90) {Real multi-writer handwriting};
  \node[stage,fill=green!8] (g) at (0,-3.48) {Compatible realizations already exist};
  \node[stage,fill=green!8] (h) at (0,-4.06) {Generation $\rightarrow$ matching $\rightarrow$ virtual writer};
  \foreach \u/\v in {a/b,b/c,c/d,d/e,e/f,f/g,g/h}{\draw[arrow] (\u) -- (\v);}
\end{tikzpicture}
\caption{Conceptual journey of the frozen project. The decisive change was the candidate space, not an additional generator module.}
\label{fig:journey}
\end{figure}

\section{From Generation to Matching}

\subsection{Real-human candidate sets}

Once character identity was treated as an equivalence class, the natural candidate set for a target character became
\begin{equation}
  \Hpool_c = \{x_{w_1,c},x_{w_2,c},\ldots\}\subset\Eclass_c,
  \label{eq:candidates}
\end{equation}
where each element is a real sample written by a different human writer or, where the corpus provides more than one sample, a distinct human realization. The pool is character-wise: the same writer need not be available for every character.

\subsection{The decisive observation}

When dozens of real realizations of the same character were displayed together, the qualitative result changed. The desired USER-compatible realization often already existed in the pool. The generator had not finally learned a hidden USER manifold. The solution was already present as a valid human sample.

This observation changes the optimization object to
\begin{equation}
  x_c^\star = \arg\min_{x\in\Hpool_c} d_u(x,c),
  \label{eq:matching}
\end{equation}
where $d_u$ measures compatibility with the USER reference set. At row level, the choice is made jointly enough to preserve consistency across characters, but each selected glyph remains an actual human realization.

\subsection{Why this is not another generator}

The previous search space was a general synthesis space, in which a model could emit pixels that were neither observed nor guaranteed to be natural. The final search space is the restricted set $\Hpool_c\subset\Eclass_c$. This restriction provides real human ink, real irregularity, and corpus-backed character identity. It does not solve arbitrary synthesis; it turns the immediate problem into compatibility selection.

The distinction is operational. A generator draws a new image. The frozen system retrieves a legal character realization and places it in a row. It may produce a new combination at the row level, but it does not claim to produce novel USER handwriting pixels.

\subsection{Coverage condition}

The reformulation has a necessary condition:
\begin{equation}
  \Hpool_c\neq\varnothing.
  \label{eq:coverage}
\end{equation}
If the real-human corpus has no candidate for $c$, matching cannot return one. The final experiments achieved complete coverage on the 197-character USER overlap and on the 100-character evaluation subset, but this is not evidence of coverage for an unrestricted Chinese character set.

\section{Virtual Writers}

The first matching baseline was not enough. Selecting the nearest physical writer assumes that one person explains the target style across the entire row. The provenance audits rejected that assumption for the strongest qualitative rows.

Table~\ref{tab:provenance} reports two anonymized whole-row examples. In the first, 32 unique writers contributed selected glyphs, with an effective writer count of 19.04. The most frequent writer contributed 20\% of the row, the top three contributed 40\%, and the top five contributed 56\%. In the second, 29 writers contributed, with effective count 18.39 and top-1, top-3, and top-5 shares of 20\%, 40\%, and 54\%. The two rows shared 25 writers. These are not the statistics of a single nearest writer.

\begin{table}[t]
\caption{Anonymized provenance of two strongly USER-like whole rows.}
\label{tab:provenance}
\centering
\scriptsize
\begin{tabular}{@{}lrrrrr@{}}
\toprule
Row & Unique & Effective & Top-1 & Top-3 & Top-5\\
 & writers & writers & share & share & share\\
\midrule
Example A & 32 & 19.04 & 20\% & 40\% & 56\%\\
Example B & 29 & 18.39 & 20\% & 40\% & 54\%\\
\bottomrule
\end{tabular}
\end{table}

The appropriate object is a virtual writer. Let the selected row be
\begin{equation}
  X^\star = \{x_{c_1}^\star,\ldots,x_{c_n}^\star\},
  \label{eq:virtual_writer}
\end{equation}
where different characters can come from different physical writers while the row is perceived as one coherent handwriting system. In this case,
\begin{equation}
  \text{physical writer identity} \neq \text{perceived handwriting style}.
  \label{eq:writer_style}
\end{equation}

The USER's direct assessment was row-level: every character in the two strong rows looked highly USER-like, and the USER could realistically have written the entire row. That evidence matters because isolated glyph hits do not establish a coherent style. The cross-writer provenance shows why a virtual writer is a better description than a hidden single-writer identity.

\section{Population-Relative Whole-Row Matching}

The final frozen method uses only real CASIA glyphs at output time. It consists of a character-wise population coordinate followed by a simple greedy row-selection rule.

\subsection{Real-ink features}

Each glyph is represented by the frozen 42-dimensional real-ink descriptor. The descriptor summarizes width and aspect, ink density, edge roughness, orientation, curvature, mass distribution, and additional morphology statistics measured directly from the real glyph. No feature is a residual against a standard font. The representation is deliberately modest: it supplies a common coordinate for comparison without pretending to be a generative model of handwriting.

\subsection{Population-relative representation}

For a glyph $x$ of character $c$, the key coordinate is not the difference from a standard raster. It is the location of $x$ within the human realization distribution for that same character:
\begin{equation}
  r(x,c)=\operatorname{Percentile}\left(x\mid P_{\mathrm{human}}(\cdot\mid c)\right).
  \label{eq:percentile}
\end{equation}
Thus the question is not ``does the USER write $c$ 12\% narrower than a standard font?'' It is ``does the USER tend to occupy the narrower part of the human realization distribution for $c$?'' When the character changes, the population coordinate is recomputed for the new character. This removes the privileged canonical reference while retaining character-specific variation.

\subsection{Matching cost}

The frozen unary cost combines the population-relative error and a normalized raw-feature error:
\begin{equation}
  d = 0.72\,d_{\mathrm{percentile}} + 0.28\,d_{\mathrm{raw}}.
  \label{eq:cost}
\end{equation}
The first term carries most of the meaning because it compares a USER reference with the real-human distribution for the same character. The raw term retains useful absolute information such as density and occupied extent.

\subsection{Candidate pruning and whole-row selection}

For each target character, the system retains the $K=20$ lowest-cost real candidates. It processes target characters in hardest-first order, so characters with the least favorable local match establish the row context before easier choices are made. A greedy whole-row selection then combines individual candidate compatibility with compatibility to the running mean signature of the selected row. The frozen row-consistency strength is 0.55.

This is an engineering selection rule, not a claim of a new optimization theory. It is simple enough to audit: compare character-wise population-relative costs, keep the top 20, process the difficult targets first, and select candidates whose running row signature remains compatible with the USER reference signature.

\subsection{What the final system does not use}

The production output is 100\% real CASIA glyphs. It does not use a generator, standard or printed font, canonical glyph, canonical skeleton, canonical contour, canonical signed-distance field, warp, blend, synthetic glyph, or standard-relative residual. It also does not call FontDiffuser at output time, although FontDiffuser is a representative example of the canonical/image-to-image generation family considered in the earlier formulation~\cite{yang2024fontdiffuser}. The final system selects and composes; it does not draw.

\begin{figure}[t]
\centering
\begin{tikzpicture}[x=1cm,y=1cm,>=Latex,
  box/.style={draw,rounded corners,align=center,minimum width=0.35\columnwidth,minimum height=0.64cm,font=\scriptsize},
  arrow/.style={->,semithick}]
  \node[box,fill=orange!10] (a) at (-2.15,0) {USER-like appearance};
  \node[box,fill=red!8] (b) at (-2.15,-1.15) {Identity-defining stroke lost};
  \draw[arrow] (a.south) -- node[right,font=\tiny,align=left,inner sep=1pt]{style pressure} (b.north);
  \node[box,fill=blue!8] (c) at (2.15,0) {Structure correct};
  \node[box,fill=gray!12] (d) at (2.15,-1.15) {Canonical-looking output};
  \draw[arrow] (c.south) -- node[left,font=\tiny,align=right,inner sep=1pt]{safety pressure} (d.north);
  \node[draw,rounded corners,fill=yellow!12,align=center,minimum width=0.84\columnwidth,minimum height=0.62cm,font=\footnotesize] (e) at (0,-2.85) {Correctness and likeness must be evaluated separately};
  \draw[arrow] (b.south) -- ++(0,-0.38) -| (e.north west);
  \draw[arrow] (d.south) -- ++(0,-0.38) -| (e.north east);
\end{tikzpicture}
\caption{Schematic summary of the two recurring early failure modes. A style-positive output could lose structure, while a safe output could remain canonical-looking.}
\label{fig:failure_modes}
\end{figure}
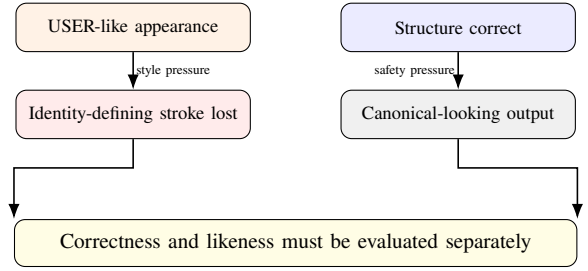

\section{Evaluation}

\subsection{Data and coverage}

The target case and the candidate corpus are summarized in Table~\ref{tab:data}. CASIA is a real multi-writer handwriting resource; the final paper reports only derived counts and selected, privacy-limited visual evidence. The offline and online counts below follow the final evaluation scope. CASIA's role as a benchmark resource is documented in the database literature~\cite{liu2013casia}.

\begin{table*}[t]
\caption{Data scope used in the frozen evaluation record.}
\label{tab:data}
\centering
\scriptsize
\begin{tabular}{@{}lrrrrl@{}}
\toprule
Source & Writers & Samples & Characters & Writer--character cells & Use\\
\midrule
USER & 1 & 200 & 197 unique & -- & references and held-out targets\\
CASIA offline & 84 & 18,662 & 207 & 17,153 & main real-human candidate pool\\
CASIA online & 84 & 12,510 & 137 & 11,504 & provenance / corpus audit\\
\bottomrule
\end{tabular}
\end{table*}

The USER/CASIA-offline character overlap was 197/197. The 100-character evaluation subset had real-human candidate coverage of 100/100. Among overlapping target characters, the mean number of one-sample-per-writer candidates was approximately 82.89. These figures establish coverage for the evaluated case; they do not establish unrestricted coverage of Chinese characters.

\subsection{Known-character held-out protocol}

The strongest comparison used a known USER character while preserving a genuine held-out protocol. For each episode, the target USER glyph was hidden. The target character was removed from the USER reference set, the hidden target image was not included in candidate ranking, and it was not included in whole-row construction. The system completed the virtual row first. Only after construction was the result compared with the genuine hidden USER glyph.

This separation is important because a retrieval system can otherwise appear successful by returning the answer image or by calibrating its candidate choice against the target. The final provenance and assembly audit found target/reference disjointness, no hidden USER record in the candidate row, and no use of the hidden target before post-hoc evaluation.

\subsection{Human evaluation}

USER likeness was judged primarily by direct USER inspection. Two qualitative levels were used. A B-level result means: ``the entire row looks like something the USER could naturally have written.'' Differences can be explained by a different pen, writing pressure, fatigue, an unstable hand, a different writing state, or normal day-to-day variation. It is clearly within the same USER handwriting system.

An A-level result means that, when compared against genuine held-out USER handwriting, the virtual result is visually almost indistinguishable. This is intentionally stricter than saying that a few glyphs look similar. It is a whole-row criterion.

\subsection{Strong whole-row regimes}

Two anonymized whole-row regimes were especially informative. The USER judged every character in each row to be highly USER-like and judged that the USER could realistically have written the entire row. These were not isolated hits: the positive assessment held across a finite row. The provenance statistics in Section~5 show that these rows were composed from dozens of physical writers, which is why they are evidence for virtual-writer coherence rather than nearest-writer identity.

\begin{figure}[t]
\centering
\includegraphics[width=\columnwidth]{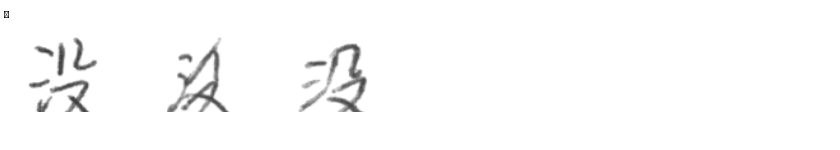}
\caption{A small crop from the real-human candidate evidence. Multiple valid realizations of the same character differ in slant, pressure, extent, and local execution while retaining character identity.}
\label{fig:equivalence}
\end{figure}

\begin{figure}[t]
\centering
\includegraphics[width=\columnwidth]{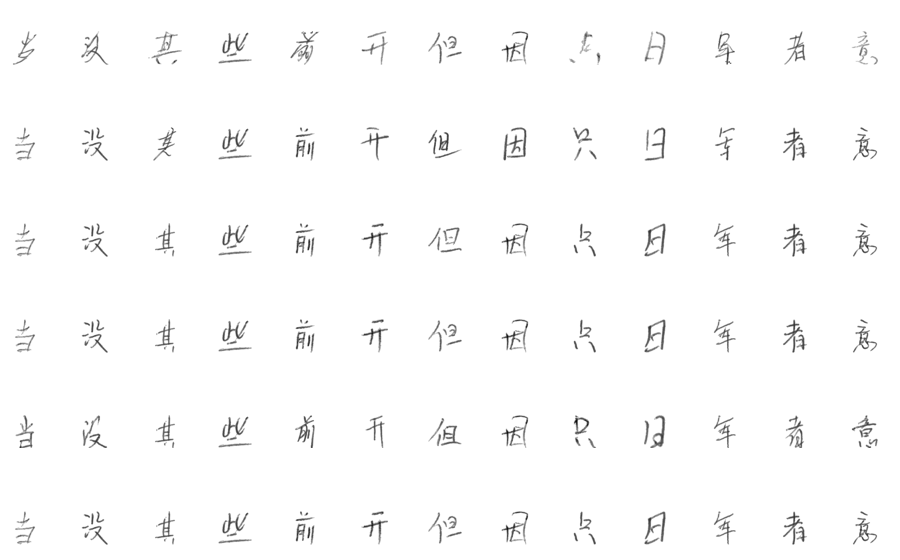}
\caption{Privacy-limited anonymous writer-row crop. Whole-row USER likeness was observed even though the selected characters came from many physical writers; the displayed row labels are anonymous review labels.}
\label{fig:virtual}
\end{figure}

\subsection{Known USER real versus virtual}

\begin{figure}[t]
\centering
\includegraphics[width=\columnwidth]{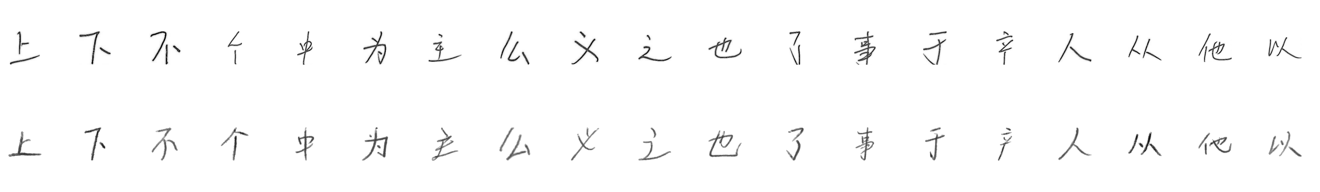}
\caption{Privacy-limited crop of the known-character held-out real-versus-virtual sheet. Only a finite subset of columns is shown; the complete USER inventory and full row are withheld.}
\label{fig:heldout}
\end{figure}

The strongest direct artifact is a limited crop of a known-character held-out real-versus-virtual row (Fig.~\ref{fig:heldout}). The USER described the comparison as visually almost indistinguishable and nearly identical by direct visual inspection. The reproduction audit found reproduced-sheet MAE equal to 0 and a different-pixel ratio equal to 0 for the virtual cells. Target and reference characters were disjoint; the hidden USER target was not used as a reference, was not used in candidate ranking, and was not used in virtual-row construction. The audit therefore rules out leakage, accidental target replacement, and row-assembly mismatch for this artifact.

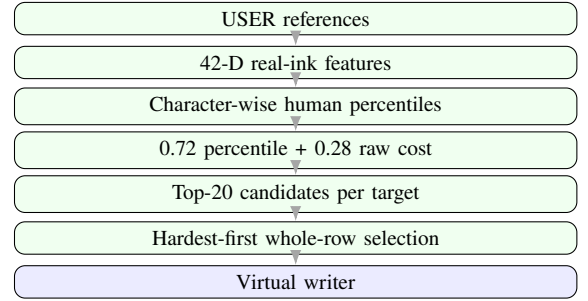
\begin{figure}[t]
\centering
\begin{tikzpicture}[x=1cm,y=1cm,>=Latex,
  pipebox/.style={draw,rounded corners,align=center,minimum width=0.84\columnwidth,minimum height=0.40cm,font=\footnotesize,fill=green!6},
  arrow/.style={->,semithick,gray!70}]
  \node[pipebox] (a) at (0,0) {USER references};
  \node[pipebox] (b) at (0,-0.58) {42-D real-ink features};
  \node[pipebox] (c) at (0,-1.16) {Character-wise human percentiles};
  \node[pipebox] (d) at (0,-1.74) {0.72 percentile + 0.28 raw cost};
  \node[pipebox] (e) at (0,-2.32) {Top-20 candidates per target};
  \node[pipebox] (f) at (0,-2.90) {Hardest-first whole-row selection};
  \node[pipebox,fill=blue!8] (g) at (0,-3.48) {Virtual writer};
  \foreach \u/\v in {a/b,b/c,c/d,d/e,e/f,f/g}{\draw[arrow] (\u) -- (\v);}
\end{tikzpicture}
\caption{Final frozen pipeline. All output glyphs are retrieved real-human CASIA samples; the pipeline contains no drawing or synthesis stage.}
\label{fig:pipeline}
\end{figure}

\subsection{Stability experiments}

The frozen stability audit contained 60 known-character held-out episodes: 12 target sets crossed with five requested reference sizes. Target sizes were 40, 80, and 120; requested reference sizes were 16, 32, 64, 128, and all available disjoint references. Every episode fell in a predefined A-like machine-proxy region.

That count is a machine result, not 60 independent human A-level judgments. The original strongest real-versus-virtual sheet was confirmed A-level by the USER; the new episodes were not all manually labeled one by one. The defensible overall conclusion is stable B-level performance, with many outputs approaching A-level under the USER-defined qualitative criterion.

\subsection{The machine--human gap}

The development repeatedly showed that a strong machine writer-latent uplift does not imply strong human likeness. Style scorers and writer metrics can fail, and a whole-row machine objective can miss a row that the USER strongly prefers. In the full set of 661 complete rows, the two manually favored rows ranked approximately 55 and 56 by the machine objective and did not enter its machine Pareto front.

Machine metrics were therefore used for screening, diagnostics, pruning, and stability checks. They were not treated as substitutes for direct USER judgment. The gap is not a nuisance detail: it is part of the reason the project is documented as an engineering development report rather than as a claim of a universal automatic likeness metric.

\section{Development Lessons}

Table~\ref{tab:lessons} abstracts the development into public technical stages. It intentionally omits internal release names and treats them as changes in assumption. The table is a compact summary; the main result is the change from the general synthesis space to a real-human candidate space.

\begin{table*}[t]
\caption{Public abstraction of the development path.}
\label{tab:lessons}
\centering
\scriptsize
\begin{tabularx}{\textwidth}{@{}p{0.16\textwidth}p{0.18\textwidth}p{0.22\textwidth}p{0.19\textwidth}Y@{}}
\toprule
Stage & Assumption & What was tried & Failure & Lesson\\
\midrule
Personalized generation & canonical to personalized & diffusion and adaptation & structure--style conflict & likeness and validity differ\\
Writer modeling & global writer identity & latent codes, tokens, adapters & machine--human mismatch & a global writer code is insufficient\\
Local rules & transferable geometry & local correspondences & unsafe execution & reusable local signal exists\\
Safe execution & constrain deformation & topology-preserving geometry & personalization suppressed & safety can erase identity\\
Representation search & a better representation solves it & broad feature comparison & canonical-looking outputs remain & the root assumption was wrong\\
Human equivalence classes & real human realizations & multi-writer samples & direct generation still weak & enter the real-human manifold\\
Matching & a compatible realization already exists & candidate retrieval & objective imperfect & generation can be unnecessary\\
Virtual writer & style is not one physical writer & cross-writer composition & machine ranking imperfect & physical identity is unnecessary\\
Frozen system & population-relative row consistency & whole-row matching & -- & stable usable solution\\
\bottomrule
\end{tabularx}
\end{table*}

Six lessons summarize the record. First, generation should not be assumed to be necessary merely because the desired output is unseen by the USER. If a rich human corpus already contains the needed realization, retrieval is the simpler and more faithful operation. Second, no single canonical realization should be given a privileged role without evidence that it is a meaningful center of the target distribution.

Third, character correctness and writer likeness must be measured as separate properties. A structure gate can protect identity while saying little about style, and a style score can reward a missing stroke. Fourth, machine writer similarity is not the same as human identity perception. Machine metrics remain useful, but their role must be calibrated to the human question.

Fifth, physical writer identity is not the same as perceived handwriting style. A row can be coherent even when its characters come from many people. Sixth, a sufficiently rich real-human candidate bank can turn synthesis into retrieval. This is not a theorem about handwriting in general; it is the engineering explanation supported by the present USER case and the available corpus coverage.

\section{Limitations}

\subsection{Candidate coverage}

The necessary condition in~\eqref{eq:coverage} remains a hard boundary. The evaluated 197-character overlap and the 100-character evaluation subset were fully covered, but the system does not establish coverage for an unrestricted Chinese character set. A missing character in the candidate corpus remains missing.

\subsection{Retrieval rather than unrestricted synthesis}

The final result is a row of retrieved and selected real-human glyphs. It is not a system for unrestricted novel handwriting synthesis from scratch. The virtual-writer effect comes from composition and selection, not from new USER-specific pixels being drawn.

\subsection{Single primary USER}

The strongest human conclusion comes from one primary USER case study. It demonstrates a successful change of formulation for that case; it does not provide a universal personalization benchmark or establish that the same candidate bank will work equally well for other users.

\subsection{Dataset and legal constraints}

The CASIA data remains subject to its source license and access conditions. The original CASIA samples are not redistributed by this report. The paper exposes derived counts and minimal visual evidence only. Any reproduction must obtain and use the source data under its own terms~\cite{casia_official}.

\subsection{Machine--human gap}

The machine objective is useful for screening and stability analysis but does not fully explain why particular whole rows are especially convincing to the USER. The ranks of the two strongest human rows are direct evidence of this gap. No machine score in this report should be read as a replacement for direct USER inspection.

\subsection{Privacy}

The complete USER handwriting inventory, the full-resolution sheet for the 100-character evaluation subset, the complete virtual font, font files, full-resolution recoverable character inventory, private repository, and reconstruction tooling are withheld. The figures show only the minimum evidence needed to explain the technical claims. This is a material limitation on independent visual replication, but it is necessary because the final system can produce highly similar handwriting.

\section{Conclusion}

The project began as personalized generation from a sparse USER sample. Canonical-centered routes repeatedly exposed a style--structure conflict, and larger writer representations or safer executors did not remove the conflict. Character equivalence classes removed the privileged role of the standard glyph. Once the search space was changed to real human realizations, the candidate pool revealed that USER-compatible glyphs already existed. Generation therefore collapsed into matching.

Whole-row selection then showed that character-level realizations from many physical writers could form a coherent virtual writer. A population-relative feature coordinate, top-20 pruning, and a frozen hardest-first greedy rule produced stable practical quality for the target use case: consistently B-level by the USER-defined criterion, with many outputs approaching A-level and a strongest held-out comparison judged visually almost indistinguishable. The result is intentionally narrower than a universal solution.

Before learning to generate a personalized realization, it is worth asking whether a sufficiently rich real-human corpus already contains one.

\bibliographystyle{IEEEtran}
\bibliography{references}

\end{document}